\documentclass[runningheads]{llncs}
\usepackage[T1]{fontenc}
\usepackage[utf8]{inputenc}
\usepackage{amsfonts}
\usepackage{amssymb,amsmath,bm,bbm}
\usepackage[font=small,labelfont=bf,tableposition=top]{caption}
\usepackage[table]{xcolor}
\DeclareCaptionLabelFormat{andtable}{#1~#2  \&  \tablename~\thetable}
\usepackage{graphicx, amsmath}
\usepackage{multicol, multirow, booktabs}
\usepackage{subcaption, cite}
\usepackage{geometry}   % YL: common it
\begin{document}
\title{AFIDAF: Alternating Fourier and Image Domain Adaptive Filters as an Efficient Alternative to Attention in ViTs
% \thanks{The work was partially supported by NSF grants DMS-2151235, DMS-2219904, and a Qualcomm Gift Award.}
}
%Vision Transformers}

\author{Yunling Zheng\inst{1}\orcidID{0000-0003-0242-6788} \and
Zeyi Xu\inst{1}\orcidID{0009-0004-8297-3136} \and
Fanghui Xue\inst{1}\orcidID{0000-0001-8366-9591} \and
Biao Yang\inst{1}\orcidID{0000-0003-2153-3247} \and
Jiancheng Lyu\inst{2} \and
Shuai Zhang\inst{2}\orcidID{0000-0002-3508-932X} \and
Yingyong Qi\inst{2}\orcidID{0000-0003-1430-258X} \and
Jack Xin\inst{1}\orcidID{0000-0002-6438-8476}
}
\authorrunning{Y. Zheng et al.}
\institute{University of California, Irvine \\
\email{yunliz1@uci.edu}, \email{zeyix1@uci.edu}, \email{fanghuix@uci.edu}, \email{biaoy1@uci.edu}, \email{jack.xin@uci.edu}\and
Qualcomm AI Research \\
\email{jianlyu@qti.qualcomm.com}, \email{shuazhan@qti.qualcomm.com}, \email{yingyong@qti.qualcomm.com}
}

% \author{Yunling Zheng\inst{1}\orcidID{0000-1111-2222-3333} \and
% Zeyi Xu\inst{1}\orcidID{1111-2222-3333-4444} \\
% \inst{1}\orcidID{2222-3333-4444-4455}\and
% 4th Author\inst{1}\orcidID{2222-3333-4444-5555}\\
% 5th Author\inst{2}\orcidID{3333-4444-5555-6666}\and
% 6th Author\inst{2}\orcidID{4444-5555-6666-7777}\\
% 7th Author\inst{2}\orcidID{5555-6666-7777-8888}\and
% 8th Author\inst{1}\orcidID{6666-7777-8888-9999}
% }
% %
% \authorrunning{Anonymous Authors}
% \institute{University of California, Irvine \\
% \email{anonymous1.com} \\
% %Princeton University, Princeton NJ 08544, USA \and
% %Springer Heidelberg, Tiergartenstr. 17, 69121 Heidelberg, Germany
% anonymous address 2 \\
% %\url{http://www.springer.com/gp/computer-science/lncs} 
% %\and
% %ABC Institute, Rupert-Karls-University Heidelberg, Heidelberg, Germany
% \email{anonymous2.com}
% %anonymous address 3
% }
%
\titlerunning{AFIDAF}
\maketitle              % typeset the header of the contribution
\begin{abstract}
We propose and demonstrate an alternating Fourier and image domain filtering approach for feature extraction as an efficient alternative to build a vision backbone without using the computationally intensive attention. The performance among the lightweight models reaches the state-of-the-art level on ImageNet-1K classification, and improves downstream tasks on object detection and segmentation consistently as well. Our approach also serves as a new tool to compress vision transformers (ViTs).

%The abstract should briefly summarize the contents of the paper in

\keywords{Fourier domain filtering \and Group shuffled large kernel convolution  \and Dual domain feature extraction.}
\end{abstract}

\section{Introduction}
Two mainstream computer vision (CV) networks are convolutional neural network (CNN, \cite{CNN_98}) and vision transformer (ViT, \cite{ViT2021}).  
ViTs have surpassed the performance of CNNs in 
recent years however at the expense of large model size and flops even though efficient attention is utilized \cite{Swin_2021}. To achieve high performance lightweight (LW) backbone models with parameter size around 5 million, attention free networks with low cost global mechanism to upgrade standard convolution has been a successful line of inquiry. For example, Fourier transform is a global convolution and can facilitate such a possibility as demonstrated in   
AFF network \cite{AFF_2023} lately. 
On the other hand, large convolution kernel vision networks \cite{VAN_22} approach this goal from the image domain, while hybrid LW models combine  mobile convolution and attention \cite{moat_2022}.
\medskip

The main contributions of our paper include:
\begin{itemize}
    \item Identify the lack of spatial mixing in AFFNet \cite{AFF_2023} and propose an {\it alternating Fourier and image domain adaptive filtering} (AFIDAF) proxy to attention in ViTs. The spatial filtering equips the   large kernel convolution \cite{VAN_22} with group shuffling operations for added efficiency.

    \item Show that AFIDAF improves AFF consistently on CV (ImageNet-1K classification and downstream) tasks while remaining in the LW category.

    \item Develop a hierachical AFIDAF framework based on Swin \cite{Swin_2021} for ViT compression while maintaining performance on CV tasks.
    
\end{itemize}
\medskip

The rest of the paper contains sections on related work, method, experiments and conclusions. 
\section{Related Work}
%Two mainstream vision  networks are 
%convolutional neural network (CNN,\cite{CNN_98}) and vision transformer (ViT, \cite{ViT2021}).  
\subsection{CNNs}
Convolution has been the basic operation of image feature extraction for over two decades \cite{CNN_98}, due its flexibility in adopting various kernel sizes for various receptive field of views to cover the image domain under translation invariance as well as its natural interpretation as filtering.
However, the convolution operation
uses static weights and so lacks adaptability across 
pixels in different parts of an image. 
It is also spatially local due to limitation of the kernel size. As a result, ViTs (\cite{ViT2021} and its variants), based on global attention originally designed for natural language processing (NLP) tasks \cite{vaswani2017attention}, outperformed well-known CNNs on computer vision (CV) tasks, see \cite{Swin_2021,segformer_21} among others. 
%\medskip

To improve CNNs to and over the level of benchmark ViTs \cite{Swin_2021,ViT2021}, additional functionalities have been introduced in recent years. One is large kernel approximation (LKA, \cite{VAN_22}) that leverages the strengths of both convolution and self-attention by 
including local structure (contextual) information, 
long-range dependence, and spatial-channel adaptability. 
Another line of inquiry is ConvNext where large kernel sizes and layer norm \cite{convnext_22}, and global response normalization layer (see \cite{convnextv2_23} for inter-channel feature competition) are 
utilized for enhancement. These developments are motivated by Swin transformers \cite{Swin_2021} yet at similar or larger capacities.

\subsection{ViTs}
Due to quadratic complexity of attention in ViT \cite{ViT2021}, efficient token mixing and global attention approximations have been actively studied with various ideas stemming from shifted window of Swin \cite{Swin_2021}. 
In lieu of window shifting, competitive performances have been reported on ImageNet-1K and downstream tasks by techniques such as pooling (Poolformer \cite{poolformer_22}), shuffling  (Shuffleformer \cite{shuffleTran_2021}), mixing across windows and dimensions (Mixformer \cite{mixformer_2022_CVPR}), high/low frequency global attention decomposition (Hiloformer \cite{hilo_2022}), pale shaped window attention (Paleformer, \cite{paleformer_2022}), cross shaped windown attention 
(CSwinformer \cite{CSwin_22}) among others. See also 
hybrid and unified CNN--ViT models \cite{cvt_2021,conv2former_22,graham2021levit,CoScale_21,mvit_2021,mvitv2_22,uniformer_23}.

\subsection{Fourier Transform based Vision Networks}
Fourier transform has been proposed first for NLP tasks \cite{fnet2022} and then found effective in promoting token mixing in CV for frequency domain filtering and feature extractions 
\cite{gfnet_2021,spectformer_23}. 
FFT is also a form of convolution,
though with a global kernel size and 
circular padding. Injecting adaptivity in the Fourier domain has been found useful for mimicing self-attention in ViTs, see \cite{adap_fno_22,AFF_2023}. 

\subsection{Lightweight Vision Networks}
Lightweight networks are desirable for mobile deployment and 
resource constrained applications. Separable (group) convolutions and shuffle operations are effective techniques for designing CNNs in the lightweight category, see MobileNets \cite{MNetV4_2024}, ShuffleNets \cite{shufflev1,shufflev2,autoshuffle_20} and references therein among others. 
Lightweight ViTs have been proposed combining MobileNet and efficient attention blocks in \cite{ViTMo_2023_ICCV,mobile_vit_22,moat_2022}, see also \cite{wang2024repvit} for a ViT motivated mobile CNN. A lightweight Fourier transform based attention-free vision network is AFF \cite{AFF_2023} which forms the baseline of our work here. 

\section{Method}
We first review the adaptive Fourier filters 
for efficient token mixing proposed in AFF \cite{AFF_2023}, point out its limitation (or lack of action in the frequency/image domain) and present our method as an alternating dual domain adaptive filter to enhance performance on visual tasks while keeping the model size in the lightweight range.

\subsection{AFF Block and Limitation}

% \begin{figure}[ht]
%     \centering
%     \includegraphics{figures/affnet_origin.pdf}
%     \caption{Caption}
%     \label{fig:affnet_origin}
% \end{figure}

% \begin{figure}[ht]
%     \centering
%     \includegraphics{figures/affnet_real.pdf}
%     \caption{Caption}
%     \label{fig:affnet_real}
% \end{figure}

\begin{figure}[ht]
    \centering
    \begin{subfigure}[t]{0.49\textwidth}
        \centering
        \includegraphics[width=\linewidth]{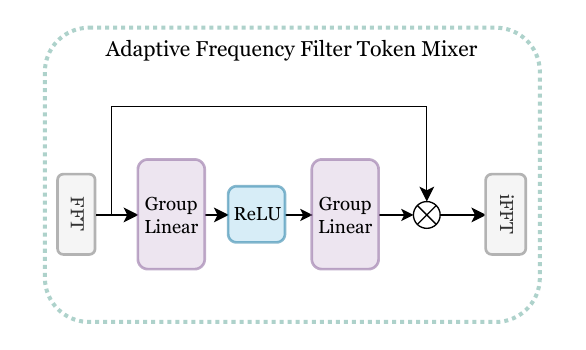}
        \caption{General (theoretical) adaptive Fourier filter proposed in AFFNet \cite{AFF_2023} for mixing features in the Fourier domain.}
        \label{fig:affnet_origin}
    \end{subfigure}
    \hfill
    \begin{subfigure}[t]{0.49\textwidth}
        \centering
        \includegraphics[width=\linewidth]{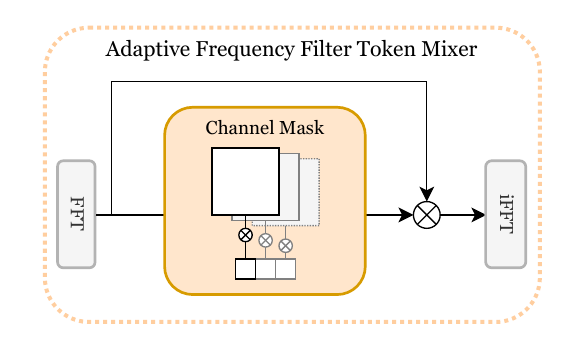}
        \caption{Implemented AFFNet, applying the Fourier domain filter channel-wise as a mask, limiting its ability to represent frequency features effectively.}
        \label{fig:affnet_real}
    \end{subfigure}
    \caption{Comparative illustration of the AFFNet block's theoretical framework (a) and its practical application (b), highlighting the discrepancy between the conceptual design and the actual implementation.}
    \label{fig:affnet_comparison}
\end{figure}

Consider feature tensor $X \in R^{H\times W\times C}$ which is mapped from an input image, 
with spatial resolution $H\times W$ and channel number $C$. A token $x\in R^{1\times 1\times C}$ is a restriction of $X$ at a fixed spatial location. Token mixing is a key operation in evolving $X$ through a deep network. A general 
expression is:
%\begin{equation}
$x^{*}_{q} := \sum_{i \in N(x_q)}\, \omega_{i,q} \, \varphi(x_i)$, 
%\label{mixtok}
%\end{equation}
where $x^{*}_{q}$ is the transformed token, 
$N(x_q)$ is a neighborhood of $x_q$ of certain size, 
$\omega_{i,q}$ the weight matrix, 
and $\varphi(\cdot)$ is embedding function. 
This formula 
%(\ref{mixtok}) 
is an abstraction of both CNN and transformer with suitable choices of $N$, $\varphi$ and $\omega$. Towards a computationally efficient, semantically adaptive and globally reaching token mixer desirable for lightweight networks, AFF \cite{AFF_2023} proposed to 1) (global) fast Fourier transform $X$ in $(h,w)$ to $F(X)$, 2) (local and adaptive on Fourier domain) mask it nonlinearly in the point-wise sense, 3) inverse Fourier back to the image domain:
\begin{equation}
    X^* = F^{-1}[ M(F(X)) \odot F(X)], \label{adapmask}
\end{equation}
where $M(\cdot)$ 
is implemented as subnetwork consisting of a group $1 \times 1$ convolution (linear) layer, followed by a ReLU function and another group linear layer; $\odot$ 
is elementwise multiplication 
(Hadamard product). The authors argued through convolution theorem that the AFF block 
(\ref{adapmask}) is global, adaptive token mixing and is mathematically equivalent to adopting a large-size dynamic convolution kernel as the weights for token mixing. An advantage of (\ref{adapmask}) is that the resulting model is attention free, CNN based, and 
lightweight with competitive performance on ImageNet-1K, though less so on downstream or dense prediction tasks (object detection and segmentation).
%\medskip

Through checking the authors' Github codes, we found that {\it the masking function $M$ actually only acts on the channel dimension while being an identity map on the frequency plane}, which limits its performance and spatial resolution. More precisely, the actually implemented AFF block is:
\begin{equation}
    X^{*}_{aff} = F^{-1}[ M_C(F(X)) \odot F(X)], \label{affmask}
\end{equation}
where $M_C(\cdot)$ is a subnet in the channel dimension leaving frequency dimensions unchanged. 
%\medskip

The main contribution of our paper is to realize that additional spatial filtering on image (or an equivalent on frequency) domain on top of (\ref{affmask}) can improve AFF while keeping model size in the lightweight range. Instead of doing so 
in the frequency domain alone (or directly on (\ref{affmask})), we propose an {\it alternating adaptive filtering methodology between image domain and Fourier domain} (AFIDAF). Abstractly, the AFIDAF block is:
\begin{equation}
    X^{*}_{afidaf} = F^{-1}[ M_C(F(M_{I}(X)) \odot F(M_{I}(X))], \label{afidaf}
\end{equation}
where $M_I(\cdot)$ is a large kernel approximation (LKA, Fig. 4 of \cite{VAN_22}) with additional grouping and shuffling. 
The LKA \cite{VAN_22} consists of depth-wise convolutions in the $H\times W$ domain followed by a CNN type multiplicative attention.
%, batch normalization and feed-forward layer. 
To be more efficient, we further downsize $M_I$ with group convolutions and shuffle operations, see the left subplot of Fig. 2(b). The alternating strategy (\ref{afidaf}) is a {\it splitting method} to handle token mixing in all $H\times W\times C$ dimensions. One potential difficulty to find a selective mask $M$ in Eq. (\ref{adapmask}) on the frequency domain is that it must be properly localized to correspond to a large receptive field of view in the image domain by the uncertainty principle of Fourier transform. On the other hand, to resolve high frequency well, the mask must also cover the corresponding part of the frequency plane. In the AFIDAF approach (\ref{afidaf}), local and high frequency features of an image (edges/corners/textures etc) are resolved by large kernel 
convolutions inside $M_I (\cdot)$ on the image domain; the low frequency and non-local features outside of individual kernel's reach are captured by channel mixing 
%$M_C$ 
%jointly by convolutions in $M_I(\cdot)$ and 
(\ref{affmask}) 
on the Fourier domain. 
So Eq. (\ref{afidaf}) is a local-global image feature extractor.
It is an interesting problem for a future study to localize (\ref{affmask}) and decrease kernel size of $M_I (\cdot)$ (hence also localize in the image domain) to reduce AFIDAF model parameter size. We present our model design next.

\subsection{AFIDAF Architecture}

\begin{figure}[!htb]
\centering
\begin{subfigure}{\textwidth}
\includegraphics[width=\textwidth]{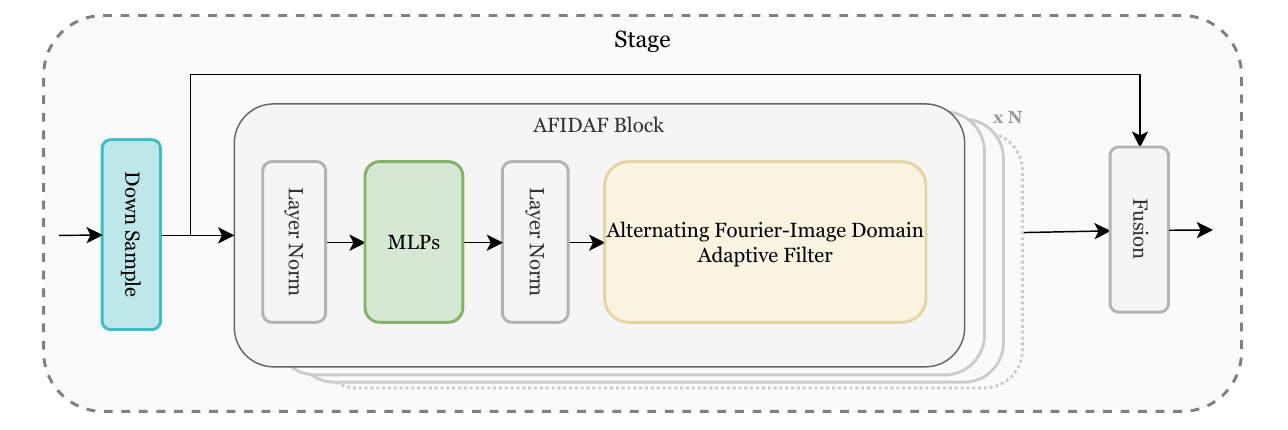}
\caption{AFIDAF block inside one of the three sequential stages of visual feature extraction in AFF architecture \cite{AFF_2023}.}
\label{fig:afidaf_stage}
\end{subfigure}
\begin{subfigure}{\textwidth}
\includegraphics[width=\textwidth]{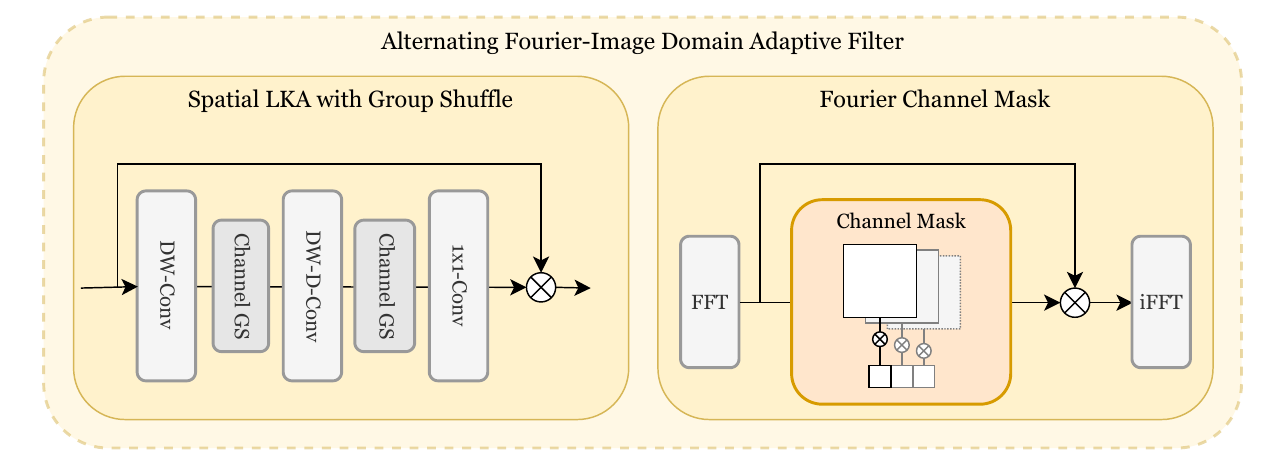}
\caption{An alternating image domain filtering (efficient large kernel convolution) and Fourier domain channelwise filtering to form a basic AFIDAF block (Fig. \ref{fig:afidaf_stage}). DW=depthwise, GS= group shuffle, DWD=depthwise-dilated.}
\label{fig:afidaf_block}
\end{subfigure}
\caption{Illustration of AFIDAF in block and stage views.}
\end{figure}

\subsubsection{Image Domain Adaptive Filtering}

To compensate for the lack of spatial filtering in AFF implementation, we propose adding a full-size kernel convolution as an adaptive filter in the image domain, prior to the Fourier domain AFF filter, then repeat this block in each of the three stages of visual feature extractions in the AFF architecture. However, employing large kernel convolution can be computationally expensive.

To mitigate the high computational cost, we implement a decomposed large kernel convolution \cite{VAN_22} combined with a channel-wise group shuffle \cite{shufflev1,autoshuffle_20}. This approach aims to reduce the computational overhead and large number of parameters typically associated with large kernel convolutions while still capturing long-range dependencies.

We adopt a convolution decomposition which includes three components: depth-wise spatial local convolution, depth-wise dilated convolution, and \(1 \times 1\) channel convolution. The depth-wise spatial local convolution focuses on proximate features, maintaining spatial locality. The depth-wise dilated convolution extends spatial coverage to capture a broader context. Finally, the \(1 \times 1\) channel convolution integrates channel-wise features, facilitating inter-channel interactions. 
%Mathematically, it is expressed as 
It is often referred to as ``attention'' in the convolutional setting (see \cite{VAN_22} and references therein). Thus, we arrive at:
\begin{equation}
    \text{Attention}_{conv} := \text{Conv}_{1 \times 1}(GS_{chan}((\text{DWD-Conv}(GS_{chan}(\text{DW-Conv}(X))))),
\label{Attention_conv}
\end{equation}
where \(X\) denotes input features, $GS_{chan}$ 
is channel-wise group shuffle.

The channel-wise group shuffle further optimizes performance by reordering the channels in each group, ensuring effective feature mixing and reducing redundancy. This step enhances the learning process by promoting diverse feature representations without significantly increasing computational costs.

By incorporating these techniques, we achieve a balanced approach that leverages both spatial and Fourier domain filters, enhancing the AFF architecture's ability to efficiently and accurately extract meaningful visual features. This approach allows us to maintain reasonable computational efficiency while achieving the desired adaptive filtering effects, see Fig. \ref{fig:afidaf_block} for the block view and Fig. \ref{fig:afidaf_stage} for the block in a stage which repeats three times from input to output.

\subsubsection{Alternating Fourier-Image Domain Filtering}

By integrating spatial and Fourier filters, we enhance the AFF architecture via a local-global approximation structure, enabling it to effectively and accurately extract significant visual features. This dual approach ensures that we maintain computational efficiency while achieving the desired adaptive filtering outcomes. Fig. \ref{fig:afidaf_block} and \ref{fig:afidaf_stage} illustrate this concept, showing the block view and its repetitive three-stage process from input to output, respectively.

\subsection{Hierarchical AFIDAF}

\begin{figure}[htb]
    \centering
    \begin{subfigure}[b]{\textwidth}
        \centering
        \includegraphics[width=\textwidth]{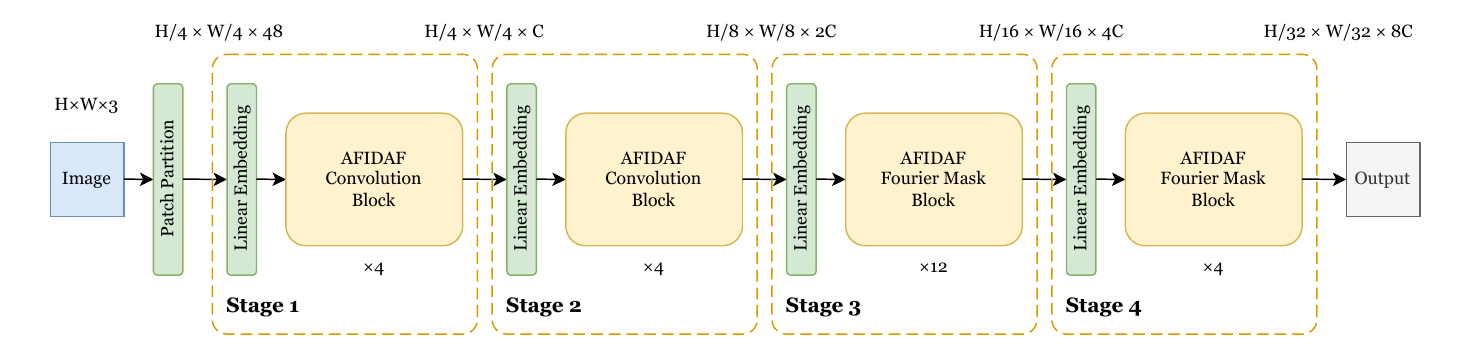}
        \caption{Hierarchical architecture of Swin \cite{Swin_2021} with its vision attention blocks replaced by AFIDAF like blocks.}
        \label{Swin-architecture}
    \end{subfigure}
    
    \begin{subfigure}[b]{\textwidth}
        \centering
        \includegraphics[scale=0.5]{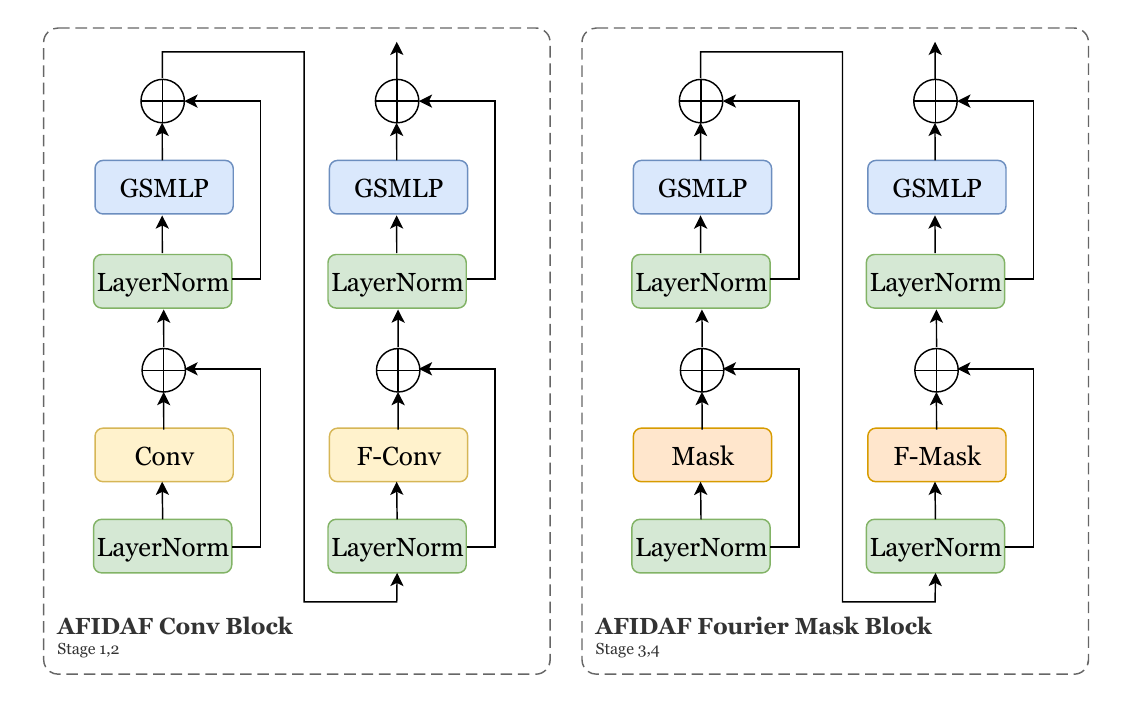}
        \caption{An alternating image filtering and Fourier channelwise mask to form a hierarchical AFIDAF block (Fig. \ref{Swin-architecture}). F-Conv=frequency domain convolution, F-Conv(X)=iFFT(Conv(FFT(X))). F-Mask=Fourier Channel Mask, F-Mask(X)=iFFT(Mask(FFT(X))), where Mask=$M_C$ as in Eq. \ref{affmask}. GSMLP is group shuffled multi-layer perception.}
        \label{Swin-block-architecture}
    \end{subfigure}
    \caption{Overview of  HAFIDAF acting on Swin \cite{Swin_2021} and the resulting compressed hierarchical architecture.}
\end{figure}

As another contribution of this paper, we improve the efficiency of existing ViTs with the dual domain alternating architecture. The Swin transformer has demonstrated a good performance with relatively low complexity among ViTs. However,  window attention computations are known to be less device-friendly than convolutions. The subsequent MLPs also rapidly increase the model size. We shall maintain the hierarchical framework of Swin (Fig. \ref{Swin-architecture}), while replacing its transformer blocks with our design of  hierarchical AFIDAF (HAFIDAF) blocks (Fig. \ref{Swin-block-architecture}).

% \begin{figure}[htb]
%     \centering
%     \includegraphics[width=\textwidth]{figures/AFIDAF_Swin.pdf}
%     \caption{Hierarchical architecture of Swin \cite{Swin_2021} with its vision attention blocks replaced by AFIDAF like blocks.}
%     \label{Swin-architecture}
% \end{figure}

% \begin{figure}[htb]
%     \centering
%     \includegraphics[scale=0.4]{figures/AFIDAF_Swin_block.pdf}
%     \caption{Hierarchical AFIDAF blocks}
%     \label{Swin-block-architecture}
% \end{figure}

\subsubsection{HAFIDAF blocks}
%Based on the 
Though in principle like AFIDAF, 
% Though sharing the same AFIDAF spirit, 
%there is a diference in 
the hierarchical AFIDAF (HAFIDAF) blocks differ in the following sense. First, HAFIDAF is for the purpose of model compression. Second, the approximations on the  spatial domain and the channel domain are made in separate stages (Fig. \ref{Swin-architecture}).

The first two stages are AFIDAF convolution blocks. Such a block consists of an alternating-type spatial/frequency convolution, which resembles the large kernel approximation $M_I$ in the setting of Fig. 2(b). Here, convolution is performed in place of window attention, as a more friendly alternative to mobile devices. Moreover, Fourier convolution is performed every 2 blocks. It has two main advantages over other simple architectures, e.g. the Hadamard product on image domain. First, F-conv acts on small frequency kernels, allowing for entries of similar frequency modes to connect. In comparison, Hadamard product acts only on single pixels. Second, compared to the Hadamard product, F-Conv (Fig. 3(b)) is smaller in size %and larger in FLOPs, and is thus 
and thus more efficient. A group shuffle MLP then follows to contribute to higher efficiency as well.

The latter two stages contain AFIDAF Fourier Mask blocks. Each block consists of the channelwise Fourier mask, which resembles the channelwise operator $M_C$ in the setting of section 3.1. 
A group shuffle MLP follows afterward.
In all stages, Layer Normalizations are performed beforehand, and shortcut connections are present for 
ease of training the deep layers.
%convenience of training optimization.

% \bf{Alternating Fourier-Image Domain Adaptive Filters}
\section{Experiments}
%We evaluate our AFIDAF model on
%several benchmarks in this section, including 
%ImageNet-1K classification, COCO object detection, and Pascal VOC segmentation.

\subsection{Image Classification}
\textbf{Setting}

The ImageNet-1K dataset \cite{russakovsky2015imagenet}, containing over 1.2 million images across 1000 object categories, is utilized for training our models from scratch to validate the effectiveness and efficiency of our proposed AFIDAF network in image classification. We trained AFIDAF from scratch for 300 epochs using $256\times 256$ pixel images on 8 NVIDIA RTX A6000 GPUs with a batch size of 1024. The learning rate schedule follows a cosine decay, starting at 2e-3 and decreasing to a minimum of 2e-4, with the AdamW optimizer (weight decay of 0.05) and cross-entropy loss.

The input features from preprocessing have a size of 256\(^2\) with 1 block and 16 output channels, passing through the network composed of 1 Conv Stem Layer and 3 Down Sample AFIDAF Blocks, concluding with the output. The Conv Stem Layer yields an output size of 64\(^2\), encompassing 4 blocks, and 32 output channels. The first Down Sample AFIDAF Block produces an output of 32\(^2\) with 2 blocks, and 96 channels for AFIDAF-T (128 for AFIDAF). The second Down Sample AFIDAF Block outputs 16\(^2\) with 4 blocks and 160 channels for AFIDAF-T (256 for AFIDAF). The third Down Sample AFIDAF Block results in an output size of 8\(^2\) with 3 blocks, and 192 channels for AFIDAF-T (320 for AFIDAF).

%\begin{table}[ht]
%\centering
%\setlength{\tabcolsep}{6pt}
%\begin{tabular}{lcccc}
%\toprule
%\multicolumn{1}{l|}{\multirow{2}{*}{Layer/Blocks}} & \multicolumn{1}{l|}{\multirow{2}{*}{Output Size}} & \multicolumn{1}{l|}{\multirow{2}{*}{Num. of Blocks}} & \multicolumn{2}{c}{Output Channel} \\
% \midrule
%\multicolumn{1}{l|}{} & \multicolumn{1}{l|}{} & \multicolumn{1}{l|}{} & \textbf{AFIDAF-T} & \textbf{AFIDAF} \\ 
%\midrule
%Image        & $256^2$ & 1 & 16   & 16   \\
%\midrule
%Conv Stem    & $64^2$  & 4 & 32   & 32   \\
%\midrule
%Down Sample & \multirow{2}{*}{$32^2$} & \multirow{2}{*}{2} & \multirow{2}{*}{96} & \multirow{2}{*}{128} \\
%AFIDAF Block &  &  &  &   \\
%\midrule
%Down Sample & \multirow{2}{*}{$16^2$} & \multirow{2}{*}{4} & \multirow{2}{*}{160} & \multirow{2}{*}{256} \\
%AFIDAF Block &   &  &   &   \\
%\midrule
%Down Sample & \multirow{2}{*}{$8^2$} & \multirow{2}{*}{3} & \multirow{2}{*}{192} & \multirow{2}{*}{320} \\
%AFIDAF Block &  &  &  &  \\
%\midrule
%\textbf{Parameters}   &         &   & 3.0M & 6.5M \\
%\textbf{FLOPs}        &         &   & 0.8G & 1.5G \\
%\bottomrule
%\end{tabular}
%\medskip
%\caption{Architecture details of AFIDAF and its tiny version 
%AFIDAF-T (built from AFF-T).  
%%with AFIDAF configuration details, including layer descriptions (output size, kernel size, output channels, and number of blocks(N)), FLOPs and parameters of models. The output channels for AFIDAF and AFIDAF-T models correspond to the AFFNet and AFF-T models \cite{AFF_2023}. The channelwise group shuffle operation is employed to reduce model complexity.
%}
%\label{tab:model-config}
%\end{table}

\subsubsection{Results}
\vspace{-10pt}
We compare our proposed AFIDAF model with other state-of-the-art lightweight models in Tab. \ref{tab:aff_imagenet}. Our AFIDAF demonstrates superior performance, achieving 80.9\% Top-1 accuracy with 6.5M parameters and 1.5G FLOPs, outperforming other lightweight networks of similar sizes. Additionally, AFIDAF-T achieves 77.6\% Top-1 accuracy with just 3.0M parameters and 0.8G FLOPs.

\subsubsection{Ablation on Alternating Domain Filtering}
To validate the effectiveness of our alternating Fourier and image domain filtering approach, we compare AFIDAF with AFFNet \cite{AFF_2023} and IDAF (replacing the AFF block with image domain LKA \cite{VAN_22}) on ImageNet-1K. The results, shown in the last 3 lines of Tab. \ref{tab:aff_imagenet}, demonstrate the superiority of our alternating domain approach over single-domain methods.

\begin{table}[ht]
\centering
\begin{tabular}{lccc}
\toprule
Model & Params (M) & Flops (G) & Top-1 (\%) \\
\midrule
MViT-XS \cite{mobile_vit_22} & 2.3 & 1.0 & 74.8 \\
EFormer-S0 \cite{ViTMo_2023_ICCV} & 3.5 & 0.4 & 75.7 \\
VAN-B0 \cite{VAN_22} &  4.1    &0.9         & 75.4  \\ 
EdgeNext-XS \cite{Maaz2022EdgeNeXt}    &  2.3 & 0.5 & 75.0      \\
AFFNet-T \cite{AFF_2023}        &  2.6 & 0.8 & 77.0      \\
{\bf AFIDAF-T} &  3.0 & 0.8 & \bf{77.6} \\
\midrule 
MNetv2  \cite{MNet_2017}       & 6.9        & 0.6               & 74.7           \\
ShuffleNetV2 \cite{shufflev2}  & 5.5        & 0.6               & 74.5           \\
MNetv3 \cite{MNet_2017}        & 5.4        & 0.2               & 75.2           \\
T2T-ViT  \cite{yuan2021tokenstotoken}      & 6.9                 & 1.8               & 76.5           \\
DeiT-T  \cite{touvron2021training}       & 5.7                 & 1.3               & 72.2           \\
CoaT-Lite-T  \cite{CoScale_21}  & 5.7                 & 1.6               & 77.5           \\
LeViT-128  \cite{graham2021levit}    & 9.2                 & 0.4               & 78.6           \\
GFNet-Ti  \cite{gfnet_2021}      & 7.0                 & 1.3               & 74.6           \\
Mformer  \cite{Mformer}      & 9.4                 & 0.2               & 76.7           \\
EfficientViT \cite{cai2023efficientvit}  
%& 7.9                 & 0.4               & 78.6 
&  7.8                & 0.7               & 79.1
\\
EdgeViT-XS   \cite{pan2022edgevits}  & 6.7                 & 1.1               & 77.5           \\
%MOne-S3        & 10.1                & 1.9               & 78.1           \\
MViT-S   \cite{mobile_vit_22}      & 5.6                 & 2.0               & 78.4           \\
EdgeNext-S \cite{Maaz2022EdgeNeXt}    & 5.6                 & 1.3               & 79.4           \\
MViTv2-1.0 \cite{mvitv2_22}    & 4.9                 & 1.8               & 78.1           \\
tiny-MOAT-1 \cite{moat_2022}&5.1 &1.2 &78.3 \\
% tiny-MOAT-2 \cite{moat_2022}& 9.8 & 2.3 &81.0 \\
MixFormer-B1 \cite{mixformer_2022_CVPR} &
8 & 0.7 & 78.9 \\
RepViT-M1.0 \cite{wang2024repvit} & 6.8 & 1.1 & 80.3 \\
 % IDAF v0\textsuperscript{1} & 7.2 & 1.6  & 80.6 \\
AFFNet \cite{AFF_2023} & 5.5 & 1.5 & 79.8 \\
IDAF & 6.2 & 1.4  & 80.3 \\
{\bf AFIDAF} & 6.5 & 1.5 &\bf{80.9} \\
\bottomrule
\end{tabular}
\caption{Lightweight network classification comparison on ImageNet-1K dataset. IDAF (image domain adaptive filtering only) replaces AFF block's channel mixing with image domain LKA \cite{VAN_22}.}
% \textsuperscript{2}v1 uses channel shuffle with layer combination based on LKA.\\
% \textsuperscript{3}AFIDAF: Alternating Fourier and Image Domain Adaptive Filters as an Efficient Alternative to Attention in Vision Transformers.
\label{tab:aff_imagenet}
\end{table}

\subsection{Object Detection}

\begin{table*}[htb]
\centering
\begin{minipage}[t]{0.48\textwidth}
    \centering
    \begin{tabular}{lcc}
        \toprule
        Model  & Param(M) & mAP(\%)\\
        \midrule
        AFFNet-T\cite{AFF_2023} & 3.0 & 25.3 \\
        \bf{AFIDAF-T} & 3.1 & \bf{25.4}  \\
        \midrule
        AFFNet\cite{AFF_2023}    & 5.6   & 28.4  \\
        IDAF & 5.9 & 28.2 \\
        \bf{AFIDAF} & 6.2  & \bf{30.2}  \\
        \bottomrule
    \end{tabular}
    \caption{Comparison of AFIDAF variants Object detection on MS-COCO 2017 dataset.}
    \label{tab:aff_coco}
\end{minipage}%
\hfill
\begin{minipage}[t]{0.48\textwidth}
    \centering
    \begin{tabular}{lcc}
        \toprule
        Model  &  Params (M) & mIOU(\%) \\
        \midrule
        AFFNet-T\cite{AFF_2023} & 3.5 & 77.8 \\
        MViTv2-0.75\cite{mvitv2_22} & 6.2 & 75.1  \\
        \bf{AFIDAF-T}  & 3.9 & \bf{79.6} \\
        \midrule
        AFFNet\cite{AFF_2023}  & 6.9  & 80.5  \\
        EdgeNext \cite{Maaz2022EdgeNeXt} & 6.5 & 80.2 \\
        IDAF & 7.5  & 81.1  \\
        \bf{AFIDAF}  & 7.8 & \bf{81.6} \\
        \bottomrule
    \end{tabular}
    \caption{AFIDAF variants vs. other LW backbones Semantic segmentation on PASCAL VOC 2012 dataset.}
    \label{tab:aff_voc}
\end{minipage}
\end{table*}

\subsubsection{Setting}
Experiments on object detection are conducted using the MS-COCO 2017\cite{lin2014microsoft} dataset, a widely-used benchmark for object detection, instance segmentation, and keypoint detection tasks. The dataset includes 118K training images, 5K validation images, and 20K test-dev images, covering 80 object categories annotated with bounding boxes, masks, and keypoints. The objects in this dataset are diverse and challenging, ranging from people and animals to vehicles and household items.

Following the common practice in \cite{Maaz2022EdgeNeXt, mobile_vit_22, AFF_2023}, we compare lightweight backbones, AFIDAF and AFIDAF-T, using the SSD\cite{liu2016ssd} framework. We initialize the backbone with ImageNet-1K pre-trained weights and fine-tune the entire model on MS-COCO for 200 epochs with a $320\times 320$ input resolution. The training uses a cosine learning rate scheduler with a base learning rate of $7\mathrm{e}{-4}$, a minimum learning rate of $7\mathrm{e}{-5}$, and AdamW optimizer (weight decay 0.05) with the Ssd Multibox loss function.

\subsubsection{Results}
\vspace{-5pt}
As shown in Tab. \ref{tab:aff_coco}, the detection models equipped with AFIDAF outperform other lightweight transformer-based detectors in terms of mean Average Precision (mAP). Specifically, AFIDAF surpasses the second-best AFFNet \cite{AFF_2023} by 1.8\% in mAP. Consistently, AFIDAF-T edges out AFFNet-T by 0.1\% in mAP with 0.1M more parameters. 

\subsection{Semantic Segmentation}

\subsubsection{Setting}
% \vspace{-10pt}
We perform semantic segmentation experiments on the PASCAL VOC 2012 benchmark dataset \cite{everingham2015pascal}. This dataset, widely utilized for object recognition, detection, and segmentation tasks, comprises of over 11,000 images with pixel-level annotations across 20 object categories. It presents significant challenges due to the high variability in object appearances, occlusions, and clutter. Following common practices \cite{Maaz2022EdgeNeXt}, we augment the dataset using MS-COCO 2017 \cite{lin2014microsoft}, incorporating additional annotations and data to enhance our experiments.

We use the DeepLabv3 \cite{chen2017rethinking} framework for semantic segmentation with AFIDAF and AFIDAF-T backbones. Images are resized to $512 \times 512$, and models are initialized with ImageNet-1K pretrained weights. Models are trained for 50 epochs on the VOC dataset, using a cosine learning rate scheduler with a base rate of $5\mathrm{e}{-4}$, a minimum rate of $1\mathrm{e}{-6}$, and optimizer AdamW with a weight decay of 0.05. The loss function employed is cross-entropy loss. 
% \subsubsection{Hierarchica Training}
% In Hierarchica Training, We utilize UperNet in mmseg as our base framework, For model compression

\subsubsection{Results}
\vspace{-5pt}
In Tab. \ref{tab:aff_voc}, AFIDAF demonstrates superior performance compared to other lightweight networks for semantic segmentation. Specifically, AFIDAF achieves a mean Intersection over Union (mIoU) of 81.6\%, surpassing the second-best lightweight network, AFFNet, by 1.1\%. Additionally, AFIDAF-T exceeds AFFNet-T by 1.8\% in mIoU. 

%\subsection{Ablation Study}
%The last 3 lines of Tab. %\ref{tab:aff_imagenet} 
%%\ref{tab:ablation-alternating}, we 
%shows the advantage of alternating Fourier and image domain filtering over AFFNet \cite{AFF_2023} and IDAF (i.e. replace AFF block with image domain LKA \cite{VAN_22}) on ImageNet-1K.
%classification. 

%\begin{table}[th]
%\centering
%\setlength{\tabcolsep}{6pt}
%\begin{tabular}{lllll}
%\hline
%                         & Model    & Parmas(M) & FLOPs(G) & Top-1(\%)$\downarrow$ \\
%\hline
%Alternating F\&I Filters       & AFIDAF   & 6.5       & 1.5      & 80.9                              \\
%Image (I) domain LKA & IDAF & 6.2       & 1.4      & 80.3                               \\
%Fourier (F) Channel Filter   & AFFNet   & 5.5 & 1.5      & 79.8  \\
%\hline
%\end{tabular}
%\medskip
%\caption{\textbf{Ablation Study of Filters on ImageNet-1K.} AFFNet \cite{AFF_2023} only has channel dimension filtering in the Fourier domain, IDAF adopts image domain LKA \cite{VAN_22} in lieu of AFF block, AFIDAF alternately filters in the Fourier and image domains, showing best accuracy with mild (no) parameter (flops) increase.}
%\label{tab:ablation-alternating}
%\end{table}

\begin{table*}[htb]
    \centering
    \begin{minipage}{\textwidth}
        \centering
        \begin{tabular}{lcccc}
            \toprule
            Model & Params (M) & Flops (G) & Top-1 (\%) & Top-5 (\%)\\
            \midrule
            Swin-T\cite{Swin_2021} & 28 & 4.5 & 81.2 & 95.5\\
            SpectFormer-XS\cite{spectformer_23} & 20 & 4.0  & 80.2 & 94.7\\
            PoolFormer-S12\cite{poolformer_22} & 12 & 1.8  & 77.2 & -\\
            PoolFormer-S24\cite{poolformer_22} & 21 & 3.4  & 80.3 & -\\
            GFNet-XS\cite{gfnet_2021}  & 16 & 2.9 & 78.6 & 94.2\\
            \textbf{HAFIDAF} & 14.8 & 4.45 & 79.8 & 95.0\\
            \bottomrule
        \end{tabular}
        \caption{Comparison of HAFIDAF with middleweight networks on ImageNet-1K classification. HAFIDAF achieves competitive performance with fewer parameters.}
        \label{tab:img-cls-hierarchical}
    \end{minipage}

    \begin{minipage}{0.48\textwidth}
        \centering
        \begin{tabular}{lcccc}
            \toprule
            Model  & Param & AP\textsuperscript{box} & AP\rlap{\textsubscript{50}}\textsuperscript{box} & AP\rlap{\textsubscript{75}}\textsuperscript{box} \\
            & (M)   & (\%) & (\%) & (\%) \\
            \midrule
            R-50\cite{he2016deep}   & 82 & 46.3 & 64.3 & 50.5 \\
            DeiT-S\cite{touvron2021training} & 80 & 48.0 & 67.2 & 51.7 \\
            Swin-T\cite{Swin_2021} & 86 & 50.5 & 69.3 & 54.9 \\
            \textbf{HAFIDAF} & 72  & 48.9 & 67.6 & 53.4  \\
            \bottomrule
        \end{tabular}
        \caption{Object detection performance on COCO dataset. HAFIDAF maintains competitive performance with significantly fewer parameters compared to larger models.}
        \label{tab:hierarchical-afidaf-coco}
    \end{minipage}%
    \hfill
    \begin{minipage}{0.48\textwidth}
        \centering
        \begin{tabular}{lcccc}
            \toprule
            Model  & Param & mIoU & mAcc & aAcc \\
            & (M)   & (\%) & (\%) & (\%) \\
            \midrule
            Swin-T\cite{Swin_2021} & 60 & 71.1 & 77.9 & 93.4 \\
            \textbf{HAFIDAF} & \textbf{46}  & \textbf{72.4} & \textbf{80.3} & \textbf{93.8}  \\
            \bottomrule
        \end{tabular}
        \caption{Semantic segmentation performance on Pascal VOC 2012 dataset. HAFIDAF outperforms Swin-T baseline across all metrics with 24\% fewer parameters.}
        \label{tab:hierarchical-afidaf-voc}
    \end{minipage}
\end{table*}

\subsection{Experimental Evaluation of HAFIDAF}
We conduct experiments to evaluate our proposed Hierarchical AFIDAF (HAFIDAF) model, comparing it with state-of-the-art vision transformers models across image classification, semantic segmentation, and object detection tasks.

\subsubsection{Image Classification}
% \vspace{-5pt}
Table \ref{tab:img-cls-hierarchical} compares HAFIDAF with other middleweight networks on ImageNet-1K. Based on the Swin-T architecture, HAFIDAF reduces parameters by 47\% (14.8M vs. 28M) while only decreasing Top-1 and Top-5 accuracy by 1.4\% and 0.5\%, respectively. This demonstrates HAFIDAF's efficiency in balancing model size and performance against recent Vision Transformers.

\subsubsection{Object Detection}
% \vspace{-5pt}
We evaluate HAFIDAF using the Cascade Mask R-CNN framework (Tab. \ref{tab:hierarchical-afidaf-coco}). With consistent training settings across models, HAFIDAF achieves a 17\% reduction in model size compared to Swin-T, with only a 1.6\% drop in AP\textsuperscript{box}. This showcases HAFIDAF's ability to balance compression and accuracy in detection tasks.

\subsubsection{Semantic Segmentation}
% \vspace{-5pt}
Using the UperNet framework, we compare HAFIDAF and Swin-T on the Pascal VOC dataset (Tab. \ref{tab:hierarchical-afidaf-voc}). HAFIDAF reduces parameters by 24\% while improving all three accuracy metrics (mIoU, mAcc, and aAcc), highlighting its effectiveness in dense prediction tasks.

% (resizing the input such that the shorter side is between 480 and 800 while the longer side is at most 1333), AdamW optimizer initial learning rate of $1\mathrm{e}{-4}$, weight decay of 0.05, batch size of 16, 3x schedule (36 epochs), and ImageNet-1K pre-trained model as initialization.

% The hierarchical AFIDAF model, derived from Swin-T and further compressed, significantly reduces the parameters (from 28M to 14.8M) and FLOPs (from 4.5G to 4.45G) off the original Swin-T, while maintaining competitive performance vs. recent ViTs of similar/larger sizes, with a 1.4 (0.5) \% decrease in Top-1 (5) accuracy (refer to Tab. \ref{tab:img-cls-hierarchical}).

% In Tab. (\ref{tab:hierarchical-afidaf-coco}), HAFIDAF compresses Swin-T by 17\% in parameter size while keeping accuracy drop at 1.5\% in AP (box).

% The last 3 lines of Tab. \ref{tab:aff_imagenet} 
% %\ref{tab:ablation-alternating}, we 
% show the advantage of alternating Fourier and image domain filtering over AFFNet \cite{AFF_2023} and IDAF (i.e. replacing the AFF block with image domain LKA \cite{VAN_22}) on ImageNet-1K.

% In Tab. \ref{tab:hierarchical-afidaf-voc}, HAFIDAF compresses Swin-T by 24\% and improves all three measures of accuracy. These results highlight the effectiveness of our proposed method for dense prediction tasks.

\section{Conclusion}
We found that the channel direction filtering in AFFNet limited its performance and proposed to alternate an efficient image domain large kernel convolution approximation with AFFNet block. The dual domain feature extraction approach (AFIDAF) and its tiny version AFIDAT-T achieved consistent improvements over AFFNet and other state of the art lightweight networks in classification and downstream CV tasks. The  hierarchical version HAFIDAT successfully compressed ViT benchmark Swin-T \cite{Swin_2021}, reducing parameter size while  maintaining 
performance in similar CV tasks.

\subsubsection*{\ackname}
% \vspace{-10pt}
The work was partially supported by NSF grants DMS-2151235, DMS-2219904, and a Qualcomm Gift Award.

% \clearpage
\bibliographystyle{splncs04}
\bibliography{sections/references}
\end{document}